\documentclass[sigconf,natbib=true]{acmart}

\setcopyright{rightsretained}
\copyrightyear{2026}
\acmYear{2026}
\acmConference[AgentSearch '26]{The First Workshop on Indexing, Retrieval, and Ranking of AI Agents}{July 24, 2026}{Melbourne, Australia}
\acmISBN{}
\acmDOI{}
\renewcommand\footnotetextcopyrightpermission[1]{}

\usepackage{booktabs}
\usepackage{microtype}
\usepackage{xcolor}
\usepackage{tikz}
\usetikzlibrary{positioning,arrows.meta,fit,shapes.geometric,shadows}
\usepackage{enumitem}
\usepackage{pifont}
\usepackage[capitalize,noabbrev]{cleveref}
\newcommand{\cmark}{\ding{51}}
\newcommand{\xmark}{\ding{55}}

\title{Representation Affects Retrieval:\\A Case Study of Skill Discovery and Routing in a Multimodal Agent Harness}

\author{Kevin Dela Rosa}
\affiliation{%
  \institution{Cloudglue}
  \country{USA}
}
\email{kdr@cloudglue.dev}


\begin{document}

\begin{abstract}
A production agent harness must discover and rank, from a growing library of skills, the one most appropriate for a user's task. At small scale this selection happens \emph{in context}: the LLM planner chooses among skill representations exposed in its system prompt, without an explicit embedding-based retrieval step. We treat this in-context selection as the small-N counterpart to embedding-based skill retrieval at scale, and present a case study of how Tinycloud, a production multimodal video agent harness, represents its skills for the planner. The harness ships skills under two recurring representations: \emph{tool-skills} that wrap a single external API or system tool and serve as primitive vocabulary, and \emph{workflow-skills} that orchestrate tool-skill calls plus a template render to produce one named deliverable. The harness exposes them via two surfaces in the system prompt: an inlined-body surface (full instructions, scripts, templates) for autoloaded skills, and a one-line listing for on-demand skills. A six-task selection ablation across three exposure regimes (all-on, default, all-off) shows that full autoload selects the gold skill on every task; all-off slows execution and produces hard discovery failures; and the production default misroutes one task because its lexical signal collides with an autoloaded tool-skill that pulls planner attention away from a listed workflow-skill. The headline finding is that in-prompt exposure of skills is not monotonically helpful: partial exposure can create lexical competition that suppresses correct selection. We connect this small-N observation to recent retrieval-based skill-routing work at large scale, and frame this contribution as a case study rather than a benchmark.
\end{abstract}

\maketitle

\section{Introduction}
\label{sec:intro}

As agent harnesses ship more skills, every harness becomes a small skill-search engine. When a user asks ``coach the rep in this meeting,'' the planner must find, rank, and select from among many candidate skills the one whose capability matches the task. Concurrent work \cite{skillrouter} frames retrieval-based skill routing as a first-class problem at $\sim$80k-skill scale, training learned retrievers over skill descriptions and reporting that hiding the skill body causes a 31--44 percentage-point drop in routing accuracy. We ask the same question at the other end of the scale: what does skill selection look like inside a production harness with a dozen heterogeneous skills, where the design choice is not which retriever to train but which surface to expose each skill on?

We use \emph{retrieval} in the broad sense of selection among candidates. At our scale (13 skills exposed in the system prompt) there is no explicit embedding-based retrieval step; the LLM planner performs in-context selection, attending to skill representations already in the prompt. We treat this as the small-N counterpart to embedding-based skill retrieval: same problem (pick the right capability for the task), different mechanism (in-context attention vs.\ learned ranking), with implications that flow in both directions.

We study Tinycloud,\footnote{\url{https://tinycloud.sh}} a freely-available multimodal video agent harness \cite{kdr_videomcp_2025, kdr_raven_2025}, which exposes its skills to the planner via two surfaces in the system prompt: an inlined-body surface (full \texttt{SKILL.md} body, scripts, templates) for autoloaded skills, and a one-line-listing surface (name + short description, full instructions injected only when invoked) for on-demand skills. The \texttt{SKILL.md} format \cite{anthropic_skills_2025} has become a community-wide convention; \cite{agentskills_survey} provides a recent survey. The exposure-surface choice is conventionally framed as a context-budget decision; we treat it instead as a representation choice for in-context selection.

Two observations frame the case study. First, Tinycloud's skills cluster into two recurring categories that map cleanly onto the two exposure surfaces:
\begin{itemize}[leftmargin=*]
\item \textbf{Tool-skills} wrap a single external API or system tool and serve as primitive vocabulary; they are autoloaded (inlined-body), so the planner always has them available.
\item \textbf{Workflow-skills} orchestrate multiple tool-skill calls plus a template render to produce one named deliverable; they are on-demand (listing-only), and their full body is injected only when the user requests their specific deliverable.
\end{itemize}

\noindent Second, this representation choice has measurable effects on selection. We run a six-task selection ablation across three exposure regimes (all-on, default, all-off), measuring tool-selection correctness, wall-clock cost, and tool-call count per cell. The ablation surfaces a non-monotonicity: partial exposure (the production default) can route \emph{worse} than either full exposure or none, because an autoloaded tool-skill can lexically dominate a listed workflow-skill on prompts whose surface form matches both.

\paragraph{Contributions.}
This paper is a case study at the small-N end of the agent-search spectrum:
\begin{itemize}[leftmargin=*,topsep=2pt,itemsep=1pt]
\item A taxonomy of skills-as-representations in a production multimodal agent (\cref{sec:system}), with template count as a useful structural marker (\cref{sec:catalogue}, \cref{tab:catalogue}).
\item A six-task selection ablation (\cref{sec:empirical}) measuring how exposure surface affects in-context selection accuracy, latency, and tool-call cost.
\item An observation that in-prompt exposure is not monotonically helpful (\cref{sec:discussion}): we describe the lexical-competition failure mode, present a minimal falsification attempt, and connect the finding to retrieval-based skill-routing at large scale \cite{skillrouter, kdr_modaroute_2025}.
\end{itemize}
\noindent We frame this as a case study, not a benchmark or scaling study; the empirical floor is one harness with thirteen skills and six fixture tasks. The harness source and cataloged skills are available at \url{https://tinycloud.sh}.

\section{Background}
\label{sec:related}

\paragraph{Agent and tool retrieval.}
Skill and tool retrieval is the closest concurrent line of work. SkillRouter \cite{skillrouter} learns retrieval-based routing over $\sim$80k skills and reports the same body-vs-listing asymmetry we observe at the within-harness scale. ToolLLM \cite{toolllm} catalogues over 16k real-world APIs as agent tools and argues for tool selection at scale; Toolformer \cite{toolformer} establishes the foundational tool-use pattern; ReAct \cite{react} provides the canonical interleaved reasoning-and-acting loop. SkillsBench \cite{skillsbench} benchmarks skill efficacy across domains. Multimodal-retrieval routing more broadly is studied in \cite{kdr_modaroute_2025}.

\paragraph{Skill format, composition, and memory.}
The \texttt{SKILL.md} convention and the autoload mechanism trace to Anthropic's Agent Skills launch \cite{anthropic_skills_2025}, adopted across multiple harnesses including Claude Code \cite{claude_code}; \cite{agentskills_survey} provides a recent survey. Voyager \cite{voyager} framed skills as composable code chunks discovered via automatic curriculum; our workflow-skills resemble these but are author-curated, deliverable-shaped, and template-emitting. MemGPT \cite{memgpt} frames the context window as managed memory; the autoload flag is a static, design-time analog of its runtime paging.

\section{The Skill System}
\label{sec:system}

\subsection{Format and loader}
A skill in Tinycloud is a directory containing three artifacts: a \texttt{SKILL.md} file with YAML frontmatter (\texttt{name}, \texttt{description}, optional \texttt{autoload}; default true) and a free-form body the agent reads as procedural knowledge; an optional \texttt{scripts/} directory with executable \texttt{.ts} or \texttt{.sh} files invoked via the agent's bash tool; and an optional \texttt{templates/} directory with reference files (HTML, JSON schemas, Markdown) the agent reads then customizes.

\subsection{Surface area in the system prompt}
The autoload flag controls how a skill is surfaced to the agent at session start. Skills with \texttt{autoload: true} appear in a \emph{Skills} block of the system prompt with full per-skill detail (name, description, paths to scripts, paths to templates, plus per-skill usage guidelines maintained by the harness). Skills with \texttt{autoload: false} appear in a separate \emph{Additional Skills} block with name and description only; their full instructions are injected on demand when the user (or agent) issues \texttt{/skills:<name>}. The asymmetry is deliberate: full skill bodies and per-skill guidelines can be hundreds to thousands of tokens each, so loading every skill at every turn would consume context budget on workflows that fire only a small subset.

\subsection{Two emergent categories}
In production, the autoload split appears to correspond to a structural distinction in what each skill does:

\begin{itemize}[leftmargin=*]
\item \textbf{Tool-skills} are atomic primitives that wrap an external API or system tool: speech transcription, semantic search, file management, structured extraction, ffmpeg-style media manipulation. The agent treats them as a vocabulary that composes. They are autoloaded so the agent always has them available.
\item \textbf{Workflow-skills} are domain-specific deliverable recipes that orchestrate multiple tool-skills, an LLM reasoning step, and a template render to produce something a user explicitly asked for: a meeting breakdown, a sales-coaching report, a blog post, a YouTube publishing package. They are loaded on-demand because each is long, narrow, and fires only when its specific deliverable is requested.
\end{itemize}

\subsection{Templates as a structural fingerprint}
In our harness the \texttt{templates/} directory serves as a useful category marker alongside the \texttt{autoload} flag. Workflow-skills consistently contain \emph{exactly one} template: the deliverable shape (an HTML report, a publishing-metadata page, a blog-post layout). Tool-skills contain either zero templates or, in the special case of a primitive-library tool-skill, multiple templates that \emph{other} skills consume via a template-name flag. Thus the same artifact (\texttt{templates/}) plays two distinct roles (deliverable shape vs primitive library), and the per-skill template count helps distinguish them.

\begin{figure}[t]
\centering
\footnotesize
\resizebox{\columnwidth}{!}{%
\begin{tikzpicture}[
  node distance=4mm and 3mm,
  every node/.style={font=\footnotesize},
  box/.style={draw, rounded corners=1.5pt, align=center, inner sep=3pt, minimum height=5mm},
  user/.style={box, fill=gray!8, minimum width=44mm},
  agent/.style={box, fill=blue!8, minimum width=44mm},
  workflow/.style={box, fill=violet!10, minimum width=58mm},
  tool/.style={box, fill=green!10, minimum width=22mm},
  outbox/.style={box, fill=orange!12, minimum width=44mm},
  arr/.style={->, >=Stealth, thick},
]
\node[user] (u) {User: ``meeting breakdown for X.mp4''};
\node[agent, below=of u] (a) {Agent (LLM planner)};
\node[workflow, below=of a] (w) {\textbf{Workflow-skill:} \texttt{meeting-breakdown/generate}};
\node[tool, below left=5mm and -8mm of w] (d) {VLM describe \\ (per-segment)};
\node[tool, below=5mm of w] (e) {VLM extract \\ (custom schemas)};
\node[tool, below right=5mm and -8mm of w] (r) {Template render \\ (HTML)};
\node[outbox, below=15mm of e] (o) {HTML deliverable (artifact)};

\draw[arr] (u) -- (a);
\draw[arr] (a) -- node[right, font=\scriptsize]{\texttt{bash} + \texttt{/skills:meeting-breakdown}} (w);
\draw[arr] (w) -- (d);
\draw[arr] (w) -- (e);
\draw[arr] (w) -- (r);
\draw[arr] (d.south) -- (o.north west);
\draw[arr] (e) -- (o);
\draw[arr] (r.south) -- (o.north east);
\end{tikzpicture}%
}
\caption{A workflow-skill (\textbf{purple}) orchestrates tool-skill calls (\textbf{green}): one or more VLM describe / extract jobs plus a template render, producing one user-facing deliverable (\textbf{orange}). The agent invokes the workflow-skill once via bash; the underlying tool-skills are orchestrated inside the skill's orchestration script, not step-by-step by the agent. This frames the workflow-skill as a recipe over a tool-skill vocabulary.}
\label{fig:sequence}
\end{figure}
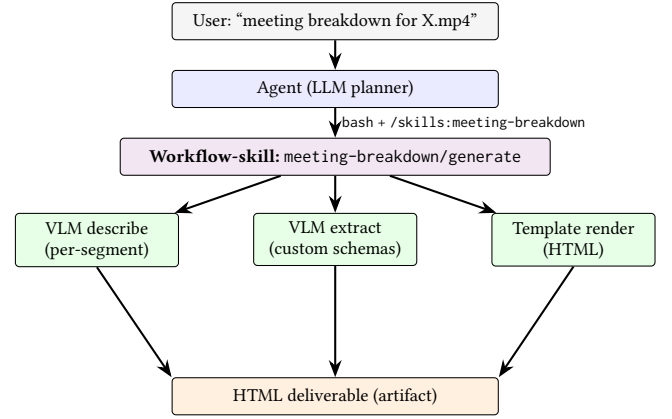

\cref{fig:sequence} shows how a workflow-skill orchestrates tool-skills under the hood for a single user request.

\section{Catalogue}
\label{sec:catalogue}

\cref{tab:catalogue} lists the twelve skills currently shipped with the harness, classified along the tool / workflow distinction. The autoload column is the production default; \emph{scripts} counts non-utility scripts (excluding shared helpers); \emph{templates} counts files in \texttt{templates/}. In this harness, the split is consistent: every workflow-skill is on-demand and contains exactly one template; every tool-skill is autoloaded and contains either zero templates or, in one case, a primitive library of templates that other skills consume.

\begin{table}[h]
\caption{The twelve production skills. \emph{Cat.} is T (tool-skill) or W (workflow-skill); \emph{Auto.} is the production autoload flag; \emph{Sc/Tp} = (non-utility scripts) / (template files). Workflow-skill templates shape the deliverable; \texttt{media-artifact}'s seven templates are a primitive library consumed by other skills.}
\label{tab:catalogue}
\centering
\scriptsize
\setlength{\tabcolsep}{3pt}
\begin{tabular}{llccl}
\toprule
Skill & Cat. & Auto. & Sc/Tp & Role \\
\midrule
\multicolumn{5}{l}{\textit{Tool-skills (atomic primitives)}} \\
\midrule
\texttt{video-analyze}      & T & \cmark & 7/0 & VLM describe + extract \\
\texttt{cloudglue-files}   & T & \cmark & 7/0 & Fetch / describe by ID \\
\texttt{data-connectors}    & T & \cmark & 6/0 & Connector ingestion \\
\texttt{collection-search}  & T & \cmark & 4/0 & Semantic search + Q\&A \\
\texttt{captions}           & T & \cmark & 2/0 & SRT / VTT / transcripts \\
\texttt{local-search}       & T & \cmark & 1/0 & Grep cached enrichments \\
\texttt{media-artifact}     & T & \cmark & 2/7 & Render template library \\
\midrule
\multicolumn{5}{l}{\textit{Workflow-skills (deliverable recipes)}} \\
\midrule
\texttt{meeting-breakdown}  & W & \xmark & 2/1 & Timeline + chapters \\
\texttt{sales-coaching}     & W & \xmark & 2/1 & Coaching dashboard \\
\texttt{ad-analysis}        & W & \xmark & 2/1 & Hook + pacing + CTA \\
\texttt{blog-post}          & W & \xmark & 2/1 & Long-form post \\
\texttt{youtube-publish}    & W & \xmark & 2/1 & Title / desc / chapters \\
\bottomrule
\end{tabular}
\end{table}

Two patterns are worth highlighting. Tool-skills vary in script complexity but uniformly carry no deliverable template (they are vocabulary, not output), while all five workflow-skills share an identical fingerprint --- two scripts (an orchestrator and a renderer) plus one deliverable template --- because the recipe is the same: orchestrate tool-skill calls, synthesize, render. The one tool-skill carrying templates (\texttt{media-artifact}) holds a reusable \emph{library} of seven primitives that other workflow-skills consume by name; the same artifact directory thus plays two distinct roles, and the per-skill template count helps signal which.

\section{Empirical}
\label{sec:empirical}

\paragraph{Question.}
What does the autoload split cost? We measure the system-prompt token budget under three loading conditions, decomposing the cost into (a) the static-rule overhead the harness emits regardless of skill set and (b) the marginal cost of each skill that surfaces as autoloaded.

\paragraph{Conditions.}
We compare three configurations of the \cref{tab:catalogue} skill set:
\begin{itemize}[leftmargin=*]
\item \textbf{All-on:} every skill has \texttt{autoload: true} (the full \emph{Skills} block lists every skill at full detail, including its scripts and template paths).
\item \textbf{Default:} the production configuration (tool-skills autoloaded, workflow-skills on-demand).
\item \textbf{All-off:} every skill has \texttt{autoload: false} (the entire skill set surfaces in the lightweight \emph{Additional Skills} block as name and description only; full bodies and per-skill paths inject only via \texttt{/skills:<name>}).
\end{itemize}
Two production skills sit outside the \cref{tab:catalogue} taxonomy: \texttt{yt-dlp} (a utility skill omitted as it doesn't fit either category) is included in the ablation under each condition's loading rule, while a Cloudglue SDK skill is force-autoloaded as a system invariant under all three conditions and excluded from the ablation; it contributes an additional $\sim$1{,}862 tokens of body content to every prompt regardless of condition. The ablation thus measures the marginal effect of the remaining 13 skills (the 12 catalogued plus \texttt{yt-dlp}). Token counts are computed by running Tinycloud's system-prompt builder under each condition and tokenizing the output with \texttt{tiktoken cl100k\_base}\footnote{Used here as a public-tokenizer proxy for Claude Opus's tokenizer, which is not publicly distributable. Token ratios across conditions are insensitive to the tokenizer choice.}.

\begin{table}[h]
\caption{System-prompt token budget by loading condition. \emph{Skills block} is just the \texttt{Skills} + \texttt{Additional Skills} sections; \emph{Static rules} is the fixed walkthrough and per-tool guidelines block emitted regardless of skill set.}
\label{tab:autoload}
\centering
\footnotesize
\setlength{\tabcolsep}{4pt}
\begin{tabular}{lrrr}
\toprule
& All-on & Default & All-off \\
\midrule
Skills block (tokens)        & 1{,}789 & 1{,}411 &   457 \\
Static rules + tools (tokens) & 6{,}379 & 6{,}379 & 6{,}379 \\
\midrule
Total system prompt (tokens) & 8{,}168 & 7{,}790 & 6{,}836 \\
\midrule
Skills-block ratio vs.\ All-off & 3.91$\times$ & 3.09$\times$ & 1.00$\times$ \\
Total prompt ratio vs.\ All-off & 1.19$\times$ & 1.14$\times$ & 1.00$\times$ \\
\bottomrule
\end{tabular}
\end{table}

\paragraph{Token-budget findings.}
The absolute autoload tax is modest: the entire workflow-skill bundle adds $\sim$378 tokens to the prompt vs.\ \emph{Default}, and even \emph{All-on} sits within the same order of magnitude as \emph{All-off}. The skills block itself varies $\sim$4$\times$, but static-rule overhead ($\sim$6.4k tokens) dominates the total prompt. The autoload split modestly reduces prompt surface area; the larger practical effect, we will show next, appears in routing behavior rather than raw token savings.

\paragraph{Tool selection and task success.}
We complement the budget measurement with a six-task ablation under each loading condition. Each cell is a single headless run of Tinycloud v0.2.1 (with Claude Opus 4.6 as the planner) that emits a structured JSON trace of the planner's tool calls; $n{=}1$ per condition--task pair, no paraphrase or seed sweep. The working directory contained $\sim$30 unrelated videos beyond the fixture targets. Per cell we record (a) which skill the agent invokes to do the task's work and (b) whether the canonical deliverable is produced. The fixtures: \emph{q1} ``meeting breakdown for [meeting].mp4'' (gold: \texttt{meeting-breakdown}); \emph{q2} ``coach the rep in [meeting].mp4'' (\texttt{sales-coaching}); \emph{q3} ``analyze [video].mp4 ad: hook structure, pacing, CTA'' (\texttt{ad-analysis}); \emph{q4} ``describe [clip].mp4'' (\texttt{video-analyze}); \emph{q5} ``find every moment across the 4 sales meetings where pricing or fees are discussed'' (\texttt{collection-search}); \emph{q6} ``make a video player for [meeting].mp4'' (\texttt{media-artifact}). q1--q3 target workflow-skills, q4--q6 target tool-skills. We treat this as a small in-house complement to larger skill benchmarks \cite{skillsbench}; \cref{tab:beta} summarizes the eighteen-cell result.

\begin{table}[h]
\caption{Tool selection per fixture task. \cmark/\xmark{} indicates whether the agent invoked the gold skill; the canonical deliverable was produced in every \cmark{} cell and in no \xmark{} cell. \texttt{wall-seconds/tool-calls} follows. q1--q3 target workflow-skills, q4--q6 target tool-skills.}
\label{tab:beta}
\centering
\scriptsize
\setlength{\tabcolsep}{3pt}
\begin{tabular}{lccc}
\toprule
Task & Default & All-on & All-off \\
\midrule
q1 & \cmark\ 194/6  & \cmark\ 201/7  & \cmark\ 262/11 \\
q2 & \cmark\ 223/7  & \cmark\ 206/10 & \xmark\ 58/7 \\
\textbf{q3} & \textbf{\xmark\ 231/13} & \textbf{\cmark\ 157/13} & \textbf{\cmark\ 192/12} \\
q4 & \cmark\ 93/7   & \cmark\ 101/5  & \cmark\ 147/14 \\
q5 & \cmark\ 298/24 & \cmark\ 277/16 & \xmark\ 1038/83 \\
q6 & \cmark\ 168/4  & \cmark\ 261/13 & \cmark\ 447/23 \\
\midrule
correct  & 5/6  & \textbf{6/6} & 4/6 \\
mean s   & 201  & 200          & \textbf{357} \\
mean calls & 10.2 & 10.7       & \textbf{25.0} \\
\bottomrule
\end{tabular}
\end{table}

\paragraph{Routing findings.}
Three observations. First, All-on is the only condition that hits gold-skill selection on every cell; Default misses q3 and All-off misses q2 and q5. Second, mean wall-clock is roughly tied between Default and All-on but $\sim$1.8$\times$ higher under All-off, driven by the agent's need to filesystem-search for skill scripts when no skill body is in the prompt; q5/All-off in particular escalated to subagent-delegation over raw cached describe-JSON files, costing 17 minutes and 83 tool calls. Third, q3 is the most informative cell: it fails under Default but recovers under both All-on \emph{and} All-off, indicating the autoload knob is not monotonic in helpfulness on the routing axis (\cref{sec:discussion}). With $n{=}6$ per condition, the 95\% binomial CI on per-condition routing accuracy spans $\pm 0.30$ (e.g., Default's $5/6 = 0.833 \pm 0.298$); the Default-vs-All-on difference is not statistically significant in isolation, and we rely on the qualitative q3 mechanism evidence in \cref{sec:discussion} to interpret it.

\section{Discussion}
\label{sec:discussion}

\paragraph{Prompt exposure is not monotonically helpful.}
The headline finding of the routing ablation is the q3 cell. The prompt --- ``Analyze the honda\_eleveate.mp4 ad: hook structure, pacing, CTA'' --- routes to the \texttt{ad-analysis} workflow-skill under both All-on (where its body is inlined) and All-off (where it is listed alongside everything else), but routes incorrectly to \texttt{video-analyze} under Default. Default is the only condition where \texttt{video-analyze} appears with its full body inlined while \texttt{ad-analysis} appears only as a one-line listing, so the lexical match \emph{Analyze$\rightarrow$\texttt{video-analyze}} dominates the listing-only match \emph{ad$\rightarrow$\texttt{ad-analysis}}. We use \emph{lexical competition} here to mean surface-form token overlap between the user query and an autoloaded skill's inlined body that pulls planner attention toward that skill, away from a competing skill whose name matches but whose body is absent from the prompt. Partial prompt exposure can therefore create lexical competition that hurts routing relative to either full or none. As a minimal falsification, we renamed \texttt{video-analyze}'s SKILL.md \texttt{name} field to \texttt{video-describe} (a name without the token ``Analyze'') and reran q3 under Default; the agent still routed to the renamed skill rather than \texttt{ad-analysis}, producing the same inline analysis. We read this as evidence that the lexical pull operates over the full inlined body content (including the description, which still contained ``\emph{Analyze} videos using Cloudglue API\dots'') rather than the surface skill name alone, and that skill authors cannot reliably fix this kind of failure by renaming alone. Concurrent work on retrieval-based skill routing at $\sim$80k-skill scale reports the same body-vs-listing asymmetry as a 31--44 percentage-point routing-accuracy drop \cite{skillrouter}, complementing prior multimodal routing studies \cite{kdr_modaroute_2025}. The autoload knob is a static, design-time signal --- tractable at the dozens-of-skills scale we operate in; at thousands-of-skills scale, learned retrieval-based routing becomes necessary, and lexical interference like q3's would remain a risk but could be adapted to via training signal in a way that static autoload assignments cannot. A practical implication for skill authors: if a workflow-skill's name overlaps semantically with an autoloaded tool-skill, leaving the workflow-skill on-demand may actively hurt rather than just save context budget.

\paragraph{Failure modes under All-off.}
Two of the six All-off cells fail outright in qualitatively different ways. Q2 produces no deliverable: after searching seven plausible skill locations on disk, the agent asks the user to manually run \texttt{/skills:sales-coaching}. Q5 bypasses the gold tool entirely, escalating to subagent-delegation over cached describe-JSON files at 17 minutes and 83 tool calls. The other All-off cells succeed but pay a 35--170\% wall-clock premium for filesystem search. The autoload knob's role on the routing axis is therefore less about raising the capability ceiling than about reducing the variance of how the agent gets there.

\paragraph{A practical heuristic for skill authoring.}
The pattern in \cref{tab:catalogue} suggests a practical test in this harness for whether a new skill should be authored as a tool or a workflow: \emph{does it produce a single artifact a user can name?} If yes (a meeting breakdown, a YouTube package, a blog post), we found it useful to treat it as a workflow-skill, on-demand with a deliverable template. If no (a search, an extraction, a transcoding), we found it useful to treat it as a tool-skill, autoloaded with no template. In this harness, the autoload pattern aligns with this distinction: workflow-skills fire only when the user requests the deliverable, while tool-skills are vocabulary that has to be present.

\paragraph{Composition.}
Workflow-skills in this harness are leaves: they call tool-skills, but no workflow-skill calls another. Higher-level workflows (e.g.\ ``quarterly review report from $N$ sales calls'') prefer $N$ tool-skill chains over an inner workflow-skill, since the inner deliverable shape is rarely the right shape for the outer's.

\paragraph{Generality.}
We described the split for multimodal video agents because that is the domain we have shipped. We hypothesize that a similar distinction may arise wherever a fragmented tool surface meets deliverable-oriented users (audio production, GIS, computational biology), though the present evidence is limited to one harness.

\paragraph{Why no embedding-retrieval baseline.}
At our scale (13 skills exposed in the system prompt) there is no candidate set to retrieve from, only representations the planner already attends to. Running an embedding-retrieval or learned-router baseline would not address the question this paper asks: how does the prompt-side representation of skills affect in-context selection? The right comparison points at this scale are exposure regimes (which we ablate), not retriever architectures. We expect retriever-based comparisons to become the right baselines as skill libraries grow past the size where every skill body can fit in the prompt; the q3 lexical-competition mechanism is a candidate failure mode for those systems as well.

\paragraph{Scaling future work.}
Three extensions would strengthen the present case study without changing its scope. First, does the lexical-competition mechanism persist as skill libraries grow toward the hundreds-or-thousands of skills SkillRouter \cite{skillrouter} considers, or does it dissolve into the retriever's noise floor? Second, do production user task logs (real prompts, not fixture prompts) surface the same partial-exposure failure mode at higher rates? Third, can a learned autoload policy --- treating the static design-time \texttt{autoload} flag as a learned per-prompt decision --- close the q3-style gap without paying the All-on token cost? Each of these is straightforward to instrument but out of scope for the present case study.

\bibliographystyle{ACM-Reference-Format}
\bibliography{main}

@misc{anthropic_skills_2025,
  title        = {Equipping Agents for the Real World with Agent Skills},
  author       = {{Anthropic}},
  year         = {2025},
  month        = oct,
  howpublished = {Anthropic Engineering Blog},
  note         = {\url{https://www.anthropic.com/engineering/equipping-agents-for-the-real-world-with-agent-skills}, October 16, 2025}
}

@article{voyager,
  title   = {Voyager: An Open-Ended Embodied Agent with Large Language Models},
  author  = {Wang, Guanzhi and Xie, Yuqi and Jiang, Yunfan and Mandlekar, Ajay and Xiao, Chaowei and Zhu, Yuke and Fan, Linxi and Anandkumar, Anima},
  journal = {Transactions on Machine Learning Research},
  year    = {2024}
}

@inproceedings{toolllm,
  title     = {{ToolLLM}: Facilitating Large Language Models to Master 16000+ Real-world {APIs}},
  author    = {Qin, Yujia and Liang, Shihao and Ye, Yining and Zhu, Kunlun and Yan, Lan and Lu, Yaxi and Lin, Yankai and Cong, Xin and Tang, Xiangru and Qian, Bill and others},
  booktitle = {International Conference on Learning Representations (ICLR)},
  year      = {2024}
}

@inproceedings{toolformer,
  title     = {Toolformer: Language Models Can Teach Themselves to Use Tools},
  author    = {Schick, Timo and Dwivedi-Yu, Janvier and Dess{\`\i}, Roberto and Raileanu, Roberta and Lomeli, Maria and Hambro, Eric and Zettlemoyer, Luke and Cancedda, Nicola and Scialom, Thomas},
  booktitle = {Advances in Neural Information Processing Systems (NeurIPS)},
  year      = {2023}
}

@article{memgpt,
  title   = {{MemGPT}: Towards {LLM}s as Operating Systems},
  author  = {Packer, Charles and Wooders, Sarah and Lin, Kevin and Fang, Vivian and Patil, Shishir G. and Stoica, Ion and Gonzalez, Joseph E.},
  journal = {arXiv preprint arXiv:2310.08560},
  year    = {2023}
}

@misc{claude_code,
  title        = {Claude Code},
  author       = {{Anthropic}},
  year         = {2024},
  howpublished = {Anthropic Documentation},
  note         = {\url{https://code.claude.com/docs}}
}

@inproceedings{react,
  title     = {{ReAct}: Synergizing Reasoning and Acting in Language Models},
  author    = {Yao, Shunyu and Zhao, Jeffrey and Yu, Dian and Du, Nan and Shafran, Izhak and Narasimhan, Karthik and Cao, Yuan},
  booktitle = {International Conference on Learning Representations (ICLR)},
  year      = {2023}
}

@article{agentskills_survey,
  title   = {Agent Skills for Large Language Models: Architecture, Acquisition, Security, and the Path Forward},
  author  = {Xu, Renjun and Yan, Yang},
  journal = {arXiv preprint arXiv:2602.12430},
  year    = {2026}
}

@article{skillrouter,
  title   = {{SkillRouter}: Skill Routing for {LLM} Agents at Scale},
  author  = {Zheng, YanZhao and Zhang, ZhenTao and Ma, Chao and Yu, YuanQiang and Zhu, JiHuai and Wu, Yong and Xu, Tianze and Dong, Baohua and Zhu, Hangcheng and Huang, Ruohui and Yu, Gang},
  journal = {arXiv preprint arXiv:2603.22455},
  year    = {2026}
}

@article{skillsbench,
  title   = {{SkillsBench}: Benchmarking How Well Agent Skills Work Across Diverse Tasks},
  author  = {Li, Xiangyi and Chen, Wenbo and Liu, Yimin and Zheng, Shenghan and Chen, Xiaokun and He, Yifeng and Li, Yubo and You, Bingran and Shen, Haotian and Sun, Jiankai and others},
  journal = {arXiv preprint arXiv:2602.12670},
  year    = {2026}
}

@inproceedings{kdr_videomcp_2025,
  title     = {{VideoMCP}: {MCP}-Enabled Video Intelligence for Multimodal Agent Reasoning and Enterprise Video Collection Analysis},
  author    = {Dela Rosa, Kevin and Xiao, Amy and Pua, Matt},
  booktitle = {Proceedings of the 2nd Workshop on Agentic AI for Enterprise: Emerging Applications and Real-world Challenges (KDD AAE)},
  year      = {2025},
  month     = aug,
  address   = {Toronto, ON, Canada}
}

@article{kdr_raven_2025,
  title   = {{RAVEN}: An Agentic Framework for Multimodal Entity Discovery from Large-Scale Video Collections},
  author  = {Dela Rosa, Kevin},
  journal = {arXiv preprint arXiv:2504.06272},
  year    = {2025}
}

@inproceedings{kdr_modaroute_2025,
  title     = {Smart Routing for Multimodal Video Retrieval: When to Search What},
  author    = {Dela Rosa, Kevin},
  booktitle = {Proceedings of the IEEE/CVF International Conference on Computer Vision Workshops (ICCVW)},
  year      = {2025}
}

\end{document}